\documentclass[runningheads]{llncs}

\usepackage{graphicx}

\usepackage{amsmath,amsfonts,bm}

\def\eqref#1{equation~\ref{#1}}

\def\1{\bm{1}}

\DeclareMathAlphabet{\mathsfit}{\encodingdefault}{\sfdefault}{m}{sl}
\SetMathAlphabet{\mathsfit}{bold}{\encodingdefault}{\sfdefault}{bx}{n}

\usepackage{url}

\usepackage{graphicx, color}
\usepackage[utf8]{inputenc}
\usepackage[english]{babel}
\usepackage[bottom]{footmisc}

\usepackage{tikz}
\usepackage{comment}
\usepackage{amsmath,amssymb} %
\usepackage{multirow}
\usepackage{adjustbox}
\usepackage{color}
\usepackage{ragged2e}
\usepackage{xcolor}         %
\usepackage{graphicx}
\usepackage{subfigure}
\usepackage{placeins}
\usepackage{float}
\usepackage[utf8]{inputenc} %
\usepackage[T1]{fontenc}    %
\usepackage{url}            %
\usepackage{booktabs}       %
\usepackage{amsmath}
\usepackage{amsfonts}       %
\usepackage{nicefrac}       %
\usepackage{microtype}      %
\usepackage{authblk}
\usepackage{nicefrac}       %
\usepackage{microtype}      %
\usepackage{xcolor}         %
\newcommand{\x}{\mathbf{x}}

\newcommand{\z}{\mathbf{z}}

\newcommand{\btheta}{\boldsymbol{\theta}}
\newcommand{\bphi}{\boldsymbol{\phi}}

\usepackage{amsmath}
\usepackage{amssymb}
\usepackage{mathtools}
\usepackage{wrapfig}
\usepackage{sidecap}
\usepackage{graphicx}
\usepackage{etoc}

\usepackage{colortbl} %
\definecolor{mColor1}{rgb}{0.9,0.9,0.9}
\definecolor{mColor2}{rgb}{0.95,0.95,0.95}
\definecolor{non-photoblue}{rgb}{0.64, 0.87, 0.93}
\definecolor{lightblue}{rgb}{0.81, 0.94, 1.0}
\definecolor{lightorange}{rgb}{1, 0.886, 0.682}

\usepackage{tikz}
\usepackage{pifont}

\usepackage{colortbl} %
\definecolor{mColor1}{rgb}{0.9,0.9,0.9}
\definecolor{mColor2}{rgb}{0.95,0.95,0.95}
\definecolor{non-photoblue}{rgb}{0.64, 0.87, 0.93}
\definecolor{lightblue}{rgb}{0.81, 0.94, 1.0}
\usepackage{pifont}

\usepackage{etoc}

\definecolor{mColor1}{rgb}{0.9,0.9,0.9}
\definecolor{mColor2}{rgb}{0.95,0.95,0.95}
\definecolor{non-photoblue}{rgb}{0.64, 0.87, 0.93}
\definecolor{lightblue}{rgb}{0.81, 0.94, 1.0}
\definecolor{lightorange}{rgb}{0.965, 0.835, 0.71}

\definecolor{mColor1}{rgb}{0.9,0.9,0.9}
\definecolor{mColor2}{rgb}{0.95,0.95,0.95}
\definecolor{non-photoblue}{rgb}{0.64, 0.87, 0.93}
\definecolor{lightblue}{rgb}{0.81, 0.94, 1.0}
\usepackage{tikz}
\usepackage{pifont}
\usepackage{etoc}
\usepackage{algorithm}
\usepackage{algorithmic}
\usepackage{xcolor}

\usepackage{hyperref}
\usepackage{dirtytalk}

\newcommand{\blue}[1]{\textcolor{black}{#1}}

\newcommand{\bluetext}[1]{\textcolor{black}{#1}}
\MakeRobust{\say}
\begin{document}
\title{Selective Test-Time Adaptation for Unsupervised Anomaly Detection using Neural Implicit Representations}
\titlerunning{STA-AD}
\author{Sameer Ambekar\inst{1,2}, Julia A. Schnabel\thanks{Shared last-authorship} \inst{1,2,3}, Cosmin I. Bercea$^*$ \inst{1,2}}
\authorrunning{Ambekar et al.}

\institute{School of Computation, Information and Technology, Technical University of Munich, Germany \and Institute of Machine Learning in Biomedical Imaging, Helmholtz Munich, Germany \and School of Biomedical Engineering and Imaging Sciences, King’s College London, UK}

\maketitle              %

\begin{abstract}

Deep learning models in medical imaging often encounter challenges when adapting to new clinical settings unseen during training. Test-time adaptation offers a promising approach to optimize models for these unseen domains, yet its application in anomaly detection (AD) remains largely unexplored. AD aims to efficiently identify deviations from normative distributions; however, full adaptation, including pathological shifts, may inadvertently learn the anomalies it intends to detect. We introduce a novel concept of \emph{selective} test-time adaptation that utilizes the inherent characteristics of deep pre-trained features to adapt \emph{selectively} in a zero-shot manner to any test image from an unseen domain. This approach employs a model-agnostic, lightweight multi-layer perceptron for neural implicit representations, enabling the adaptation of outputs from any reconstruction-based AD method without altering the source-trained model. Rigorous validation in brain AD demonstrated that our strategy substantially enhances detection accuracy for multiple conditions and different target distributions. Specifically, our method improves the detection rates by up to 78\% for enlarged ventricles and 24\% for edemas. Our code is available: \url{https://github.com/compai-lab/2024-miccai-adsmi-ambekar}.

\begin{figure*}[t]
\centering

\includegraphics[width=0.882\linewidth]{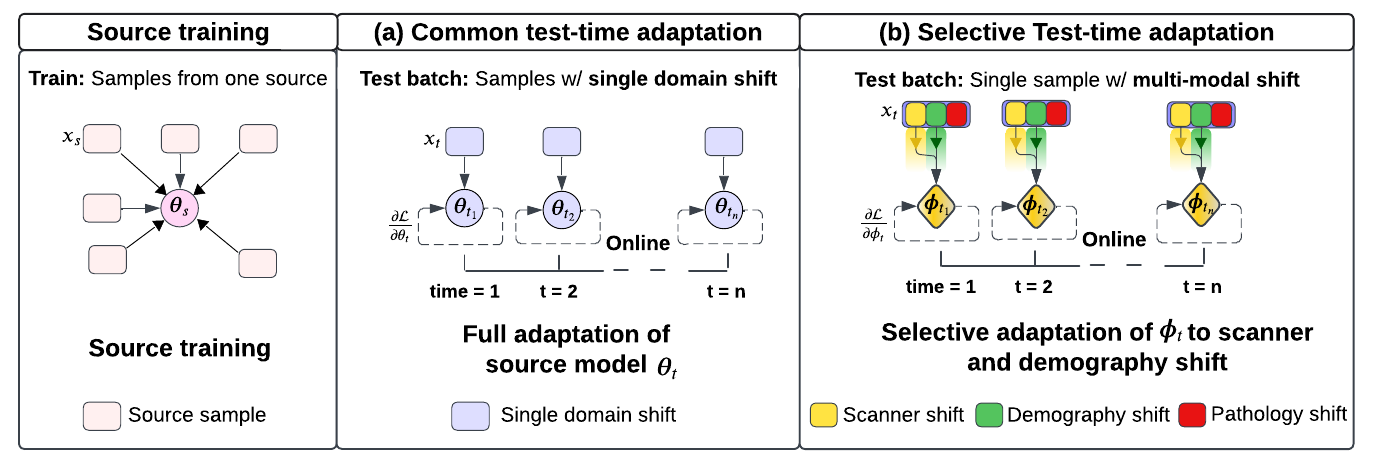}

\caption{{\emph{Selective} test-time adaptation. (a) Common test-time adaptation use the source model $\btheta_{s}$ to obtain $\btheta_{t}$ by fully adapting to $\x_{t}$ through iterative updates. (b) In contrast, we propose a new concept of \emph{selective} test-time adaptation. We learn $\bphi_{t}$ and adapt \emph{selectively} to  $\x_{t}$, implicitly excluding pathology shifts.}
}
\label{fig1:intro}

\end{figure*}

\end{abstract}

\section{Introduction}

Unsupervised anomaly detection (UAD) commonly starts by training models on large datasets comprising control patients presumed to be healthy. These models are subsequently deployed in various clinical environments to detect deviations from expected normative anatomical structures~\cite{behrendt2023patched,bercea2023generalizing,pinaya2022unsupervised,zimmerer2019unsupervised}. 

However, these models tend to underperform when deployed in new, unseen clinical environments due to various distribution changes, such as those caused by different imaging equipment, often referred to as scanner shifts~\cite{bercea2023bias,pandey2020target}. Even though it is common practice to train on datasets different from those in deployment settings, the resulting domain shifts are largely unaddressed. A common yet flawed approach to mitigate this issue involves using 2D slices from 3D pathological scans labeled for their absence of pathology~\cite{wolleb2024binary}. This approach, however, presents two critical challenges: the slices may inadvertently contain pathological traces, and there is a risk of data leakage when slices from the same patient are used to represent both healthy and pathological conditions. An alternative that seems straightforward but is complex in implementation is to fine-tune the model on the target data distribution~\cite{liang2020we,wang2021tent}. While effective, this strategy demands extensive data on control patients, which many specialized clinics do not possess~\cite{song2023ecotta,ambekar2022skdcgn}.

To address this issue, a recent approach known as \say{test-time adaptation} has emerged \cite{ambekar2023learning,iwasawa2021test,sun2020test,wang2021tent}, which is applied to adapt to target distributions affected by domain-shifts~\cite{liu2021ttt++,wang2021tent}, or in generating target datasets from a few samples~\cite{chen2022generative,ojha2021few,xiao2022few}. In this setting, the model is trained only on data from the source domain and then adapted to make predictions on unseen target data. However, a significant challenge in AD lies in adjusting for unknown variations caused by different scanners and patient demographics while preserving the sensitivity needed to detect subtle pathological changes. This necessitates a \emph{selective} adaptation approach, where the algorithm adjusts to recognize and compensate for non-pathological variations without compromising its ability to identify anomalies. Current adaptation methods~\cite{liang2020we,iwasawa2021test,wang2021tent} often struggle in these scenarios because they either fail to address multi-modal distribution shifts, or they fully adapt to target data~\cite{iwasawa2021test,liang2020we,ojha2021few,wang2021tent,xiao2022few} as shown in Figure~\ref{fig1:intro}(a).

In this paper, we introduce a novel \emph{selective} test-time adaptation framework for anomaly detection \emph{(STA-AD)}, see Figure~\ref{fig1:intro}(b). Our framework is designed to enhance existing AD methods by adapting them to new target distributions on pathological scans in a zero-shot manner.  In summary, our contributions are:
\newpage

\begin{itemize}

\item We propose the novel concept of \emph{selective} test-time adaptation for AD. To the best of our knowledge, this innovation is the first to enable AD models to adjust for domain shifts directly on pathological target data.
\item Our method leverages an efficient neural implicit learning model that bypasses the need for extensive retraining or large curated datasets, thereby considerably simplifying deployment in clinical settings.
\item We demonstrate the efficacy of our \emph{STA-AD} framework through extensive validation, resulting in considerable improvements in detection accuracy across various conditions and target distributions.

\end{itemize}

\section{Materials and methods}
\begin{figure*}[t]
\centering
\includegraphics[width=0.99\linewidth]{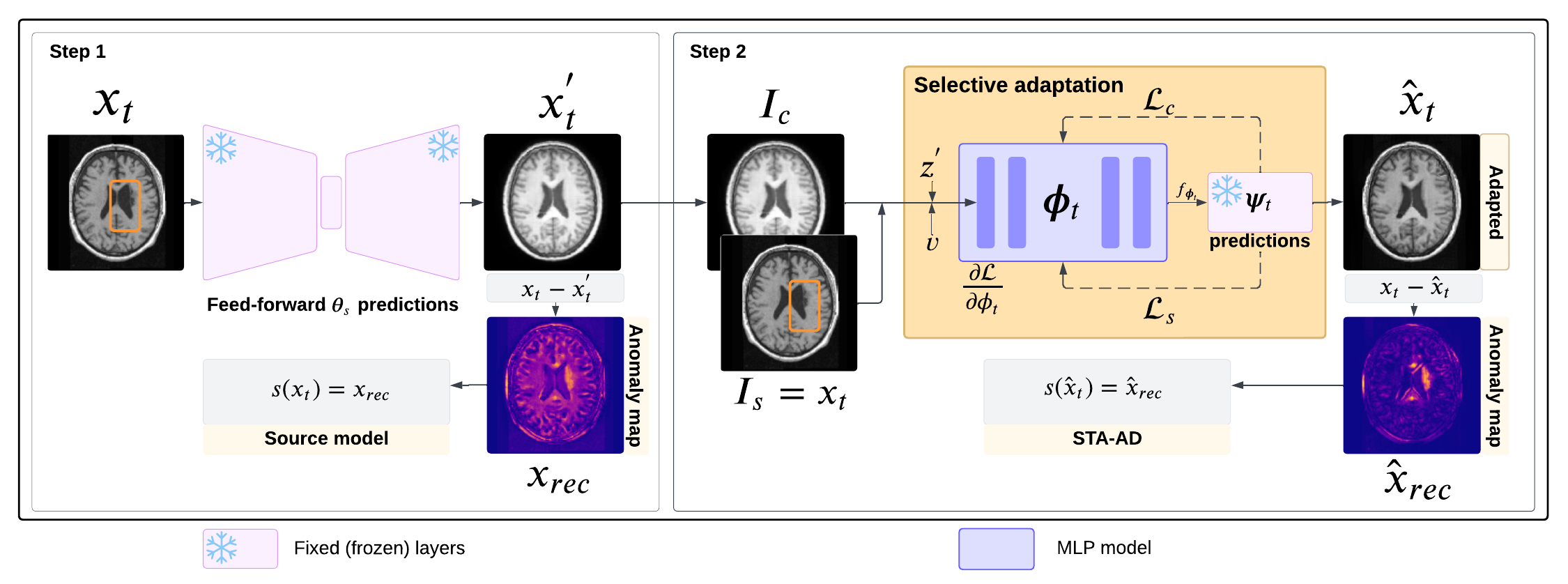}

\caption{{\textbf{{STA-AD} at test-time.} We leverage source model predictions $\x^{'}_{t}$ as \say{content} and target images $\x_{t}$ as \say{style} to train neural implicit representations $\bphi_{t}$ to adapt images $\hat{\x}_{t}$. The initial false positives due to domain shifts seen in $\x_{rec}$ are removed after adaptation ($\hat{\x}_{rec}$), while the detected lesion is highlighted.}}
\label{fig:method}
\end{figure*}

Let $\mathcal{X}$ be the space of all images $\x \in \mathbb{R}^n$, where $n$ could be 2 or 3-dimensional. AD networks are trained on a source domain $\mathcal{D}_{s} = \{(\mathbf{x}_s)^i\}_{i=1}^{N_s}$ containing a large number of control samples ($N_s$). The method can then be deployed to multiple target distributions $\mathcal{D}_{t} = \{ (\mathbf{x}_t)^i \}_{i=1}^{N_t}$.
A generative source model indicated as $\btheta_{s}$ is trained on source domains $\mathcal{D}_{s}$, by minimizing the common reconstruction losses ($\mathcal{L}_{}$). 
This process is formalized, where $\mathcal{L}$ consists of generative loss functions~\cite{bercea2023generalizing,daniel2021soft} aimed at obtaining the oracle model, represented as ${\btheta}_{s} = \arg\min_{\boldsymbol{\theta}} \mathbb{E}_{({\x}_{s}) \in \mathcal{D}_{s}} [\mathcal{L}({\x}_{s}; {\btheta})].$

If pathological shifts were the only discrepancies between the domains $\mathcal{D}_{s}$ and $\mathcal{D}_{t}$, the methods would accurately identify these deviations as anomalies. However, in reality, $\mathcal{D}_{t}$ contains multiple scanner and demographic shifts, as shown in Figure~\ref{fig1:intro}. These affect the performance of AD (see $\x_{rec}$ in Figure~\ref{fig:method}).
Furthermore, the target domain {$\mathcal{D}_{t}$ may consist solely of pathological samples, such as in specialized clinics for tumor detection. Therefore, \emph{selective} adaptation is crucial. By adjusting to non-pathological distributional shifts while disregarding pathologies, we ensure robust anomaly detection across diverse domains.\\

\noindent\textbf{Neural implicit representation learning} allows the parameterization of signals utilizing neural networks~\cite{de2023deep,grattarola2022generalised,molaei2023implicit}. %
Their compactness, resolution independence and inherent differentiability make them particularly advantageous for test-time adaptation.
In our work, we utilize them to model the adapted image function $f: \mathbb{R}^n \rightarrow \mathbb{R}^n$, where $n$ represents the dimensionality of the adapted image's flattened vector, $\hat{x}_{t}$. Here, we aim to learn a neural network $ F_{\bphi}: \mathbb{R}^n \rightarrow \mathbb{R}^n $ parameterized by weights $ \bphi $, such that $ F_{\bphi} $ approximates $ f $ as closely as possible.\\

\noindent\textbf{\emph{Selective} test-time adaptation.} 
We utilize \cite{kim2024controllable} for the test-time training. We propose to \emph{selectively} adapt to specific distribution shifts, e.g., scanner shifts with a single sample from target data. In this framework, a single target sample is treated as the style image ($I_{s}$), and its corresponding prediction from a source-trained model as the content image ($I_{c}$). By utilizing the content from the source model, we mitigate adapting to anomalies from the target domain. Using $I_{c}$ and $I_{s}$ as inputs, we train $\bphi_{t}$ at test-time to generate the target-adapted image ($\hat{\x_{t}}$), as shown in Figure~\ref{fig:method}. Initially, we generate $z_c$ and $z_s$ vectors with random normal distribution, along with a random flattened vector  $v_r$. These latent vectors are then interpolated to produce a new latent vector $z' = \text{interpolate}(z_c, z_s)$. This vector $z'$, combined with  $v_r$ serves as input for $\bphi_{t}$:
\begin{equation}
{I_{t}} = \bphi_{t}(z', \text{}(v)).
\label{Eqn:phi_pred}
\end{equation}

\blue{The $\bphi_{t}$ generates the required target-adapted image ($I_{t}$), which is then optimized using a pre-trained VGG model ($\psi$). This optimization process involves minimizing the content and style losses:}

\begin{equation}
\begin{split}
\mathcal{L}_\mathrm{cont}(\psi(\bphi_{t}(z'(i), v;\bphi_{t})),F_c) 
& + \mathcal{L}_\mathrm{style}(\psi(\bphi_{t}(z'(i), v;\bphi_{t})),F_s)
\end{split}
\label{Eqn:loss_calc}
\end{equation}
where  $\mathcal{L_{\text{cont}}} = \lVert \psi(\bphi_{t}(z',v; \bphi_{t})) - F_c \rVert^2$ is the L2 norm between the adapted image features and content features. Similarly, the style loss $\mathcal{L_{\text{style}}} = \lVert g(\psi(\bphi_{t}(z',v; \bphi))) - g(F_{s}) \rVert^2$ is the L2 norm of differences in Gram matrices ($g$) of the features from the adapted image and the style image at \blue{various layers \cite{harkonen2020ganspace}} of the $\psi$ network. 
We optimize $\bphi_{t}$ by updating it to a new state $\bphi^*_t$ through the  minimization of the content and style losses, which are weighted by $\beta$ and $\gamma$ respectively, using learning rate $\lambda{}$:
\begin{align}
\bphi^*_{t} = \bphi_{t} - \lambda \nabla_{\bphi_{t}}\big(&\beta *\mathcal{L}_\mathrm{cont}(\psi(\bphi_{t}(z', v;\bphi_{t})),F_c) \nonumber \\
&+ \gamma * \mathcal{L}_\mathrm{style}(\psi(\bphi_{t}(z', v;\bphi_{t})),F_s)\big)
\label{Eqn:tta_update}
\end{align}

Finally, we assign $\hat{\x_{t}}$ = $I_{t}$ to obtain the target-adapted image.
Note that this process optimizes deep semantic features extracted from specific layers of a pre-trained VGG model~\cite{harkonen2020ganspace}. This acts as a surrogate loss~\cite{grabocka2019learning,nguyen2009surrogate} to prevent adaptation to local/non-smooth deformations, which in our context represent pathological representations.
For all experiments, we train all the parameters of the $\bphi_{t}$ model on single test samples using Eqn.~\ref{Eqn:tta_update}. 
Finally, we compute the anomaly map between the input image ($\x_{t}$) and adapted image ($\hat{\x_{t}}$) by $s({\hat{\x}_{t}}) = |x_{t} - \hat{x_{t}}|.$

\section{Experiments and Results}
In \autoref{sec::TTA_healthy},  we explore potential distribution shifts between healthy samples from the source and target datasets and apply our approach to mitigate them.  In \autoref{sec::TTA_unhealthy}, assesses whether our \emph{selective} adaptation effectively adjusts to anomalous target samples and evaluates its impact on the downstream AD tasks. Additionally, we introduce a parameter-efficient variant of $\bphi_{t}$, which trains only 1\% of the total parameters of $\bphi_{t}$ yet delivers competitive results. Moreover, aligning with the principle that reducing entropy is a core objective of adaptation \cite{shannon1948mathematical}, we demonstrate that our adapted model minimizes entropy post-adaptation.\\

\noindent\textbf{Datasets.} \blue{We evaluate our method on public T1-weighted brain Magnetic Resonance Imaging datasets that inherently contain domain shifts due to different scanner vendors and variations in acquisition protocols. The pre-processing of the datasets follows common baselines~\cite{bercea2023generalizing}. We denote IXI~\cite{ixi_dataset} as the source dataset (S) containing healthy scans. The target dataset ($T_{+}$) is based FastMRI+~\cite{zhao2021fastmri+} containing 13 different anomaly types acquired from various vendors, reminiscent of multi-target adaptation. At test-time, we address these samples from the multiple target datasets with our single sample adaptation capabilities.}\\

\noindent\textbf{Implementation details.}  Algorithm~\ref{alg:2} outlines our approach. We train the $\bphi$ network at test time with a single sample per batch in an online manner.

\noindent We do not modify the source training process; instead, we utilize the original source model and introduce test-time learning of our MLP network. 
For our experiments, we use two anomaly detection backbones: RA \cite{bercea2023generalizing}, a generative autoencoder and DDPM \cite{ho2020denoising}, a diffusion-based model. We keep their original training protocols and freeze their model weights after the source training phase. The implementation of $\bphi$ is an MLP with a 16-layer depth and 128-unit width \cite{kim2024controllable}, and $\psi$ is a \blue{pre-trained VGG~\cite{simonyan2014very} model pre-trained on ImageNet weights.} \blue{For Table~\ref{table:metrics}, we calculate the evaluation metrics between the input image ($x_{t}$) from the test domain and our selectively adapted image ($\hat{x}_{t}$).} For Table~\ref{table:s2_metrics}, we follow \cite{bercea2023generalizing}. For the additional experiments, we utilize the same MLP with added batch normalization layers and reduce the number of parameters to be trained by only optimizing these layers. Moreover, both $\bphi_{t}$ variants are compute-friendly due to their tiny \blue{model capacity and applicable to any existing AD models. }

\begin{algorithm}[b!]
\label{alg:1}
\small
\caption{Selective test-time adaptation.\\
{\textbf{Input:}} $\mathcal{T}$: target domain; learned and frozen $\btheta_s, \bphi_t$: MLP;\\
{\textbf{Output:}} Adapted model parameters $\bphi^{*}_{t}$
}
\label{alg:2}
\begin{algorithmic}[]
\STATE Draw a single sample $\x_{t}$, set as $I_{s}$
\STATE Get $I_{c} = f_{\btheta}(x_{t})$ and initialize $\z_{c}, \z_{s},V_{r} $
\STATE From $\bphi_{t}$ generate $\hat{x}_{t}$. \textcolor{gray}{(Eqn.\ref{Eqn:phi_pred})}
\STATE Update $\bphi_{t}$ with loss from pretrained layers of $\psi$ \textcolor{gray}{(Eqn.\ref{Eqn:loss_calc})}
\STATE Optimize $ \bphi^*_{t} = \bphi_{t} - \lambda \nabla_{\bphi_{t}}(\beta *\mathcal{L}_\mathrm{cont} +  (\gamma * \mathcal{L}_\mathrm{style}) $ \textcolor{gray}{(Eqn.\ref{Eqn:tta_update})}
\STATE Set $\hat{\x_{t}} = I_{t}$; compute anomaly map $(\overline{x_{t}}) = |x_{t} - \hat{x_{t}}|$.
\end{algorithmic}
\end{algorithm}

\begin{table}[t]
\caption{\textbf{Comparisons for healthy samples from different domains}. The source model naturally performs well on source data. However, on unseen target data, the performance degrades by lower SSIM scores, higher MAE, and higher LPIPS scores. \blue{We show the best results in \textbf{bold} and \underline{underline} the runner-up.}}%
\centering
\label{table:metrics}
\resizebox{0.99\columnwidth}{!}{
    \setlength\tabcolsep{4pt} 
    \begin{tabular}{lccc}
    \toprule
    \textbf{Method} & \textbf{MAE $\downarrow$} & \textbf{SSIM $\uparrow$} & \textbf{LPIPS $\downarrow$} \\
    \midrule
    \textit{Source} & & & \\
    
    \quad RA \cite{bercea2023generalizing} & 0.060 \scriptsize{$\pm$0.005} & 0.702 \scriptsize{$\pm$0.028} & 0.049 \scriptsize{$\pm$0.007} \\ 
    \rowcolor{orange!10} 
     \quad \quad + STA-AD (ours) & \textbf{0.047} \scriptsize{$\pm$0.020} & {0.680} \scriptsize{$\pm$0.050} & \textbf{0.036} \scriptsize{$\pm$0.017} \\
    \cline{1-4}
    \quad DDPM \cite{ho2020denoising} & 0.050 \scriptsize{$\pm$0.003} & 0.753 \scriptsize{$\pm$0.024} & 0.030 \scriptsize{$\pm$0.004} \\ 
    \rowcolor{orange!10} 
     \quad \quad+ STA-AD (ours) & \textbf{0.045} \scriptsize{$\pm$0.003} & \textbf{{0.791}} \scriptsize{$\pm$0.01} & \textbf{0.021} \scriptsize{$\pm$0.002} \\
    
    \midrule
    \textit{$T_{+}$ Multi-target} & & & \\
    \quad RA \cite{bercea2023generalizing}& 0.127 \scriptsize{$\pm$0.035} & 0.465 \scriptsize{$\pm$0.077} & \underline{0.084} \scriptsize{$\pm$0.028} \\
    \quad \quad + Histogram matching & \underline{0.081} \scriptsize{$\pm$0.020} & \underline{0.535} \scriptsize{$\pm$0.069} & 0.094 \scriptsize{$\pm$0.032} \\
    \rowcolor{orange!10}
     \quad \quad + STA-AD (ours) & \textbf{0.078} \scriptsize{$\pm$0.021} & \textbf{0.600} \scriptsize{$\pm$0.070} & \textbf{0.047} \scriptsize{$\pm$0.023} \\
    \cline{1-4}
    \quad DDPM \cite{ho2020denoising} & \underline{0.080} \scriptsize{$\pm$0.008} & \underline{0.556} \scriptsize{$\pm$0.050} & \underline{0.078} \scriptsize{$\pm$0.021} \\
    \quad \quad + Histogram matching & \textbf{0.058} \scriptsize{$\pm$0.008} & \textbf{0.632} \scriptsize{$\pm$0.045} & 0.095 \scriptsize{$\pm$0.026} \\
    \rowcolor{orange!10}
     \quad \quad + STA-AD (ours) & {0.104} \scriptsize{$\pm$0.018} &{0.473} \scriptsize{$\pm$0.070} & \textbf{0.071} \scriptsize{$\pm$0.018} \\
    \bottomrule
    \end{tabular}
}
\end{table}

\begin{figure*}[tb!]
\centering
\includegraphics[width=0.98\linewidth]{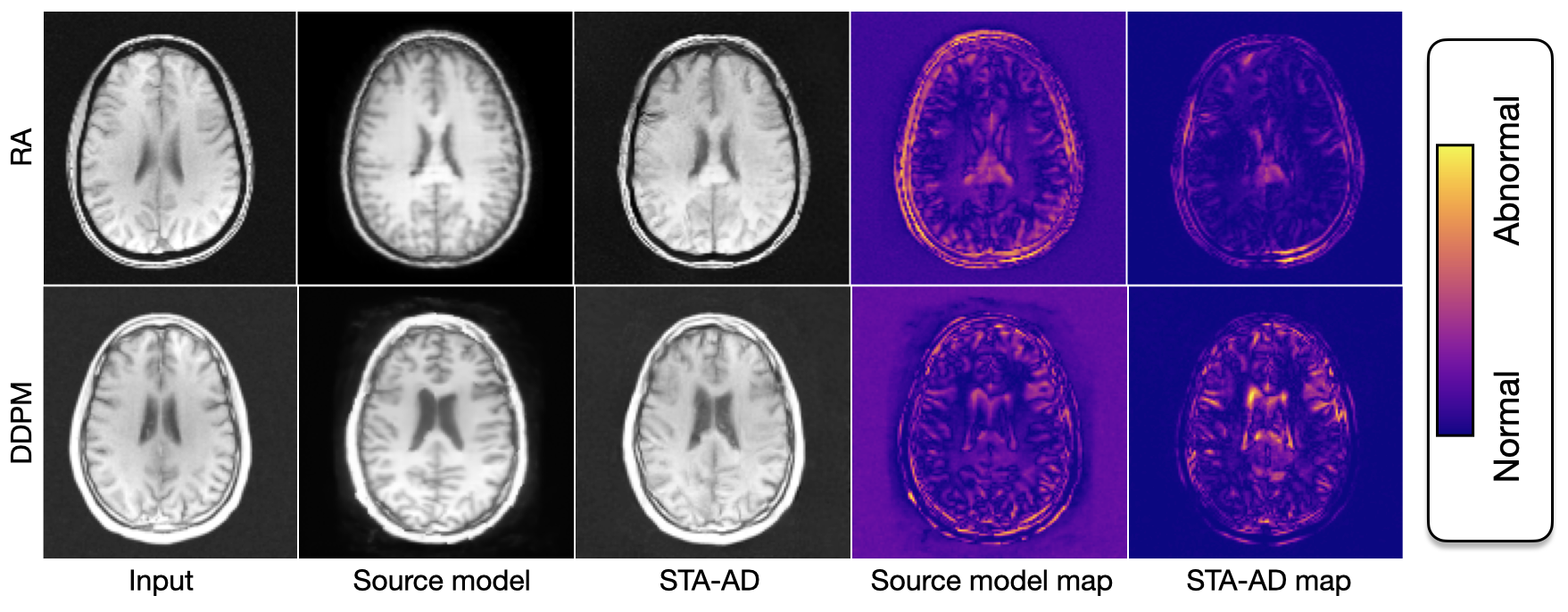}

\caption{{\textbf{Comparison on healthy samples from $T_{+}$ dataset.}} For both backbones, the source models achieve higher false positives due to inaccurate reconstructions. Our approach, \emph{STA-AD}, through the adapted image, reduces false positive detections in the presented anomaly maps.}
\label{fig:output1}
\end{figure*}
\subsection{Test-time adaptation on healthy samples\label{sec::TTA_healthy}}
We evaluate whether non-pathological distributional shifts are present between healthy samples from the source distribution and target distributions $T_{+}$ in Table~\ref{table:metrics}.
The source model performs well on source data but struggles to generalize to the target domains $T_{+}$, as evidenced by nearly halved SSIM scores and higher perceptual errors. Similarly, as shown in Figure~\ref{fig:output1}, the RA and DDPM models exhibit increased false positives in their respective anomaly maps.
\blue{As a baseline, we consider histogram matching.
This method achieves slightly higher MAE and SSIM scores than the source model but achieves lower LPIPS scores, which is also evident through the generated image artifacts provided in Supplementary. In contrast, our method consistently improves the MAE, SSIM, and LPIPS scores compared to the source model and histogram matching on both healthy source and pathological target datasets. For the DDPM backbone adapted to the target data, our method achieves improved LPIPS scores but lower MAE and SSIM values. This may be because our method intentionally preserves the original content of the source images, which might result in pixel-wise content differences. We aim to keep the source content and align the source style to the target domain.}

\subsection{Selective Test-time adaptation for anomaly detection\label{sec::TTA_unhealthy}} We evaluate the performance of our approach in detecting the various conditions from the $T_{+}$ domain in Table~\ref{table:s2_metrics} and Figure~\ref{fig:output_s2}. 
The detection task consists of 13 various conditions with one or more target images per condition. We measure the performance with true positives (TP) and F1 scores. We show the average performance (Avg.) and three exemplary conditions. The detailed results on all conditions are presented in the Supplementary.

\begin{table*}[tb!]
    \caption{\textbf{Comparisons for anomaly detection on unhealthy samples.} We shot the best results in \textbf{bold}. \textcolor{black}{{Our method outperforms both of the baselines and achieves up to 87\% improvement in average for the diffusion-based model.}}}
    \centering
    \label{table:s2_metrics}
    \setlength{\tabcolsep}{1pt}
        \begin{adjustbox}{width=0.99\columnwidth}
            \centering
            \begin{tabular}{l cc  cc  cc  cc}
                \toprule	   
                \multirow{2}{*}{\textbf{Method}} & \multicolumn{2}{c}{\textbf{Avg}.} & \multicolumn{2}{c}{\textbf{Edema}} & \multicolumn{2}{c}{\textbf{E. Ventricles}} & \multicolumn{2}{c}{\textbf{Encephalomalacia}} \\
                \cmidrule(lr){2-3} \cmidrule(lr){4-5}  \cmidrule(lr){6-7}  \cmidrule(lr){8-9} 
                & \textbf{\textbf{TP}} $\uparrow$ & {\textbf{F1} $\uparrow$}& \textbf{TP} $\uparrow$ & {\textbf{F1} $\uparrow$}& \textbf{TP} $\uparrow$ & {\textbf{F1} $\uparrow$}& \textbf{TP} $\uparrow$ & {\textbf{F1} $\uparrow$} \\\midrule
                RA \cite{bercea2023reversing} & {0.51} \scriptsize{$\pm$0.00} & 0.15 \scriptsize{$\pm$0.30}& {0.72} \scriptsize{$\pm$0.00} & 0.25 \scriptsize{$\pm$0.30} &0.47 \scriptsize{$\pm$0.30} & 0.32 \scriptsize{$\pm$0.01} & 1.00 \scriptsize{$\pm$0.00} & 0.15 \scriptsize{$\pm$0.01} \\
                \rowcolor{orange!10} 
                
                +STA-AD (ours) & \textbf{0.70} \scriptsize{$\pm$0.30} &\textbf{0.23} \scriptsize{$\pm$0.30} & \textbf{0.83} \scriptsize{$\pm$0.30} &\textbf{0.40} \scriptsize{$\pm$0.30} & \textbf{0.84} \scriptsize{$\pm$0.36} &\textbf{0.52} \scriptsize{$\pm$0.20}& 1.00 \scriptsize{$\pm$0.30} &\textbf{0.40} \scriptsize{$\pm$0.01} \\
                \midrule
                DDPM \cite{ho2020denoising} & 0.39 \scriptsize{$\pm$0.40} & 0.11 \scriptsize{$\pm$0.20}& 0.67 \scriptsize{$\pm$0.40} & 0.32 \scriptsize{$\pm$0.30} & 0.26 \scriptsize{$\pm$0.30} & 0.15 \scriptsize{$\pm$0.00} & 0.00 \scriptsize{$\pm$0.00} & 0.00 \scriptsize{$\pm$0.00} \\
                
                \rowcolor{orange!10} 
                +STA-AD (ours) & \textbf{0.73} \scriptsize{$\pm$0.40} & \textbf{0.19} \scriptsize{$\pm$0.30} &
                \textbf{0.83} \scriptsize{$\pm$0.30} & \textbf{0.33} \scriptsize{$\pm$0.20} & \textbf{0.47} \scriptsize{$\pm$0.40} & \textbf{0.25} \scriptsize{$\pm$0.30} & \textbf{1.00} \scriptsize{$\pm$0.00} & \textbf{0.29} \scriptsize{$\pm$0.00} \\
                \bottomrule
            \end{tabular}
        \end{adjustbox}
\end{table*}

\begin{figure*}[tb!]
\centering

\includegraphics[width=0.98\linewidth]{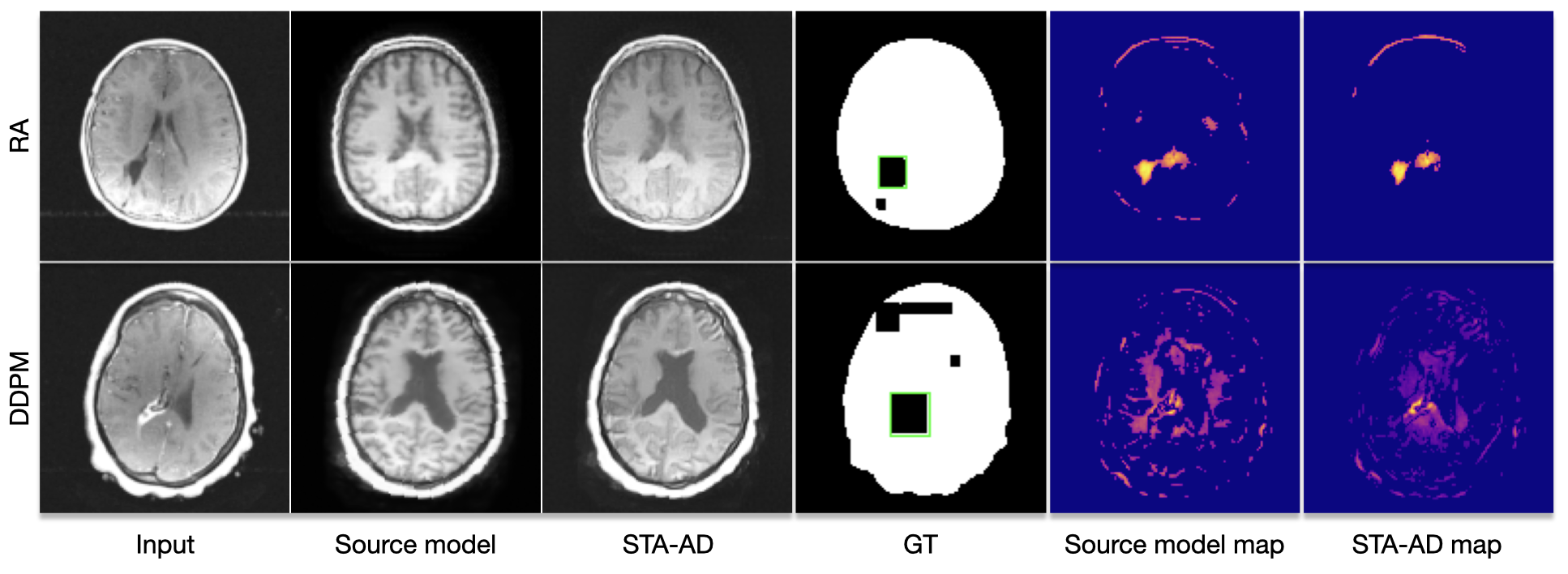}

\caption{{\textbf{Comparisons with different backbones on $T_{+}$ unhealthy samples.}} The \emph{STA-AD} framework can enhance the performance of AD methods by reducing false positives and accurately detecting anomalies after target adaptation.}
\label{fig:output_s2}
\end{figure*}

Our method demonstrates an average 87\% increase in true positives using the DDPM backbone and an average 37\% increase with the RA backbone, alongside improvements of 53\% and 72\% in overall F1 scores, respectively. Specifically, for the condition of encephalomalacia, the baseline DDPM model notably struggled, missing the anomaly. Our adapted approach successfully detected the true positive with a high accuracy of 1.0, considerably enhancing the F1 score. Similarly, for structural anomalies such as the enlarged ventricles, our method uniformly boosts true positives and F1 scores across both backbones. This pattern of increasing true positives and consistently improving F1 scores extends to other conditions as well, emphasizing our approach's effectiveness in enhancing disease detection rates and specificity through adaptation.
\bluetext{Figure~\ref{fig:output_s2} contains the visual examples. Our method significantly reduces false positives while maintaining true positive detection, such as detecting lesions close to the left posterior lateral ventricle. However, the second row shows the limitations of our approach. Although we still reduce false positive detections, the DDPM reconstruction differs too much in content from the input, resulting in suboptimal anomaly detection.} \\

\noindent \textbf{Implicit entropy minimization.} Following the principle that entropy minimization facilitates adaptation~\cite{iwasawa2021test,wang2021tent,vu2019advent,shannon1948mathematical}, we also investigate the entropy error~\cite{shannon1948mathematical} on the $T_{+}$ dataset. We report the entropy errors (lower is better) on healthy and unhealthy variants of $T_{+}$. As shown in Table~\ref{tab:ra_tsa_ad}, the predictions of source model `RA' obtain high entropy error, while our proposed approach reduces the entropy error considerably after adaptation.
By nearly halving the entropy error, our approach achieves significantly more precise adaptation.\\

\noindent \textbf{Efficiency analysis.} Our approach optimizes the $\bphi_{t}$ model with a total of 600,000 parameters in about 4 minutes per single sample. We also propose a variant that requires optimizing just ~1\% (4096 parameters) of the total parameters by adjusting the batch norm affine parameters~\cite{ioffe2015batch} and still achieving competitive performance on the $T_{+}$ dataset, as detailed in Table~\ref{tab:tta_bn}. Both proposed methods are highly efficient, requiring only one GPU for training due to their compact model sizes of 600,000 and 6,000 parameters, respectively. This demonstrates the practicality of our approach, especially in environments with limited computational resources.

\begin{table}[tb!]
\centering
\begin{minipage}{0.48\textwidth}
\centering
\caption{\textbf{Entropy minimization. }} 
\label{tab:ra_tsa_ad}
\resizebox{0.65\textwidth}{!}{%
\begin{tabular}{lcc}
\toprule
\cmidrule(lr){2-3}
& {Healthy} $\downarrow$ & {Unhealthy} $\downarrow$ \\ 
\midrule
\textbf{RA} & 13.9 & 13.9  \\
\rowcolor{orange!10}{\textbf{STA-AD}} & \textbf{6.8} & \textbf{6.7} \\
\bottomrule
\end{tabular}%
}
\end{minipage}
\hfill
\begin{minipage}{0.48\textwidth}
\centering
\caption{\textbf{Efficiency analysis.}}
\label{tab:tta_bn}
\resizebox{\textwidth}{!}{%
\begin{tabular}{llll}
\toprule
& \textbf{Parameters} & \textbf{MAE} $\downarrow$ & \textbf{LPIPS} $\downarrow$\\ 
\midrule
w/o BN & 600000  & {0.078} \scriptsize{$\pm$0.02} & {0.04} \scriptsize{$\pm$0.02}\\
w/ BN  & 4096  & 0.141 \scriptsize{$\pm$0.02} & 0.12 \scriptsize{$\pm$0.03}\\
\bottomrule
\end{tabular}%
}
\end{minipage}
\end{table}

\section{Conclusion}
In this work, we proposed a \emph{selective} adaptation paradigm, \emph{STA-AD}, that allows the adaptation of any reconstruction-based anomaly detection (AD) method directly to target pathological domains, thereby improving their performance. We validated our approach on a multi-target brain MRI dataset containing multiple lesions and demonstrated performance boosts of up to 87\%. We believe our work will encourage future research in adapting AD methods to unseen pathological distributions. Further improvements in inference times and reducing reliance on source model content predictions could enhance its clinical benefits, making it a valuable tool for deploying AD methods in various clinical settings.

\section*{Acknowledgements}
This paper is supported by the DAAD programme Konrad Zuse Schools of Excellence in Artificial Intelligence, sponsored by the Federal Ministry of Education and Research. C.I.B. is funded via the EVUK program (“Next-generation Al for Integrated Diagnostics”) of the Free State of Bavaria and partially supported by the Helmholtz Association under the joint research school ‘Munich School for Data Science’. We thank Lina Felsner for her assistance with the diagrams.

\bibliographystyle{splncs04}
\bibliography{main}

\clearpage 
\newpage

\begin{leftline}
	{
		\Large{\textsc{Appendix}}
	}
\end{leftline}

\begin{table}[h!]
\centering
\caption{\textbf{Detailed comparisons of anomaly detection on unhealthy samples.} We show the best results in \textbf{bold}. Our method consistently achieves higher True positives and higher F1 scores.}
\begin{tabular}{lcccc}
\toprule
\textbf{$T_{+}$ anomalies} & \multicolumn{2}{c}{\textbf{RA}} & \multicolumn{2}{c}{\textbf{STA-AD (Ours)}} \\
\cmidrule(lr){2-3} \cmidrule(lr){4-5}
                             & {TP mean $\uparrow$} & {F1 mean $\uparrow$} & {TP mean  $\uparrow$} & {F1 mean $\uparrow$} \\
\midrule
Absent septum pellucidum        & 0.00 & 0.00  & 0.00 & 0.00  \\
Craniatomy           & 0.47 & 0.13  & \textbf{0.67} & \textbf{0.23} \\
Dural  thickening               & 0.14 & 0.01  & \textbf{0.57} & \textbf{0.19}  \\
Edema                & 0.72 & 0.37  & \textbf{0.83} & \textbf{0.40}  \\
Encephalomalacia     & 1.00 & 0.15  & 1.00 & \textbf{0.40}  \\
Enlarged ventricles  & 0.47 & 0.32  & \textbf{0.84} & \textbf{0.52}  \\
Intraventricular     & \textbf{1.00} & 0.29  &\textbf{ 1.00} & \textbf{0.33}  \\
Lesions              & 0.55 & \textbf{0.17} & \textbf{0.64} & 0.16  \\
Posttreatment change & 0.48 & 0.11  & \textbf{0.55} & \textbf{0.12}  \\
Resection            & 0.50 & 0.26  & \textbf{0.80} & \textbf{0.38}  \\
Sinus                & 0.50 & 0.04  & \textbf{1.00} & \textbf{0.09}  \\
Wml                  & 0.20 & 0.04  & \textbf{0.40} & \textbf{0.08}  \\
Mass all             & 0.69 & 0.11  & \textbf{0.81} & \textbf{0.17}  \\
\bottomrule
\end{tabular}

\end{table}

\begin{table}[t]
\centering
\caption{\textbf{Detailed comparisons of anomaly detection on unhealthy samples.} Best results in bold. The conclusion is similar; our method performs the best. }
\begin{tabular}{lcccc}
\toprule
\textbf{$T_{+}$ conditions} & \multicolumn{2}{c}{\textbf{Diffusion}} & \multicolumn{2}{c}{\textbf{STA-AD (Ours)}} \\
\cmidrule(lr){2-3} \cmidrule(lr){4-5}
                             & {TP mean $\uparrow$} & {F1 mean $\uparrow$} & {TP mean  $\uparrow$} & {F1 mean $\uparrow$} \\
\midrule
Absent septum pellucidum       & 0.00 & 0.00  & 0.00 & 0.00  \\
Craniatomy           & 0.47 & 0.12  & \textbf{0.80} & \textbf{0.18} \\
Dural thickening               & 0.14 & 0.02  & \textbf{0.71} & \textbf{0.28}  \\
Edema                & 0.67 & 0.32  & \textbf{0.83} & \textbf{0.33}  \\
Encephalomalacia     & 0.00 & 0.00  & \textbf{1.00} & \textbf{0.29}  \\
Enlarged ventricles  & 0.26 & 0.15  & \textbf{0.47} & \textbf{0.25}  \\
Intraventricular substance     & \textbf{1.00} & 0.40  & \textbf{1.00} & \textbf{0.50}  \\
Lesions              & 0.55 & 0.14  & \textbf{0.68} & \textbf{0.18}  \\
Posttreatment change & 0.55 & 0.10  & \textbf{0.66} & \textbf{0.16}  \\
Resection            & 0.50 & 0.08  & \textbf{0.80} & \textbf{0.14}  \\
Sinus opacification                & 0.00 & 0.00  & \textbf{1.00} & \textbf{0.08}  \\
White matter lesions                  & 0.20 & 0.04  & \textbf{0.60} & \textbf{0.06}  \\
Mass             & 0.73 & \textbf{0.15 } & \textbf{0.92} & \textbf{0.15}  \\
\bottomrule
\end{tabular}

\end{table}

\begin{figure*}[b!]
\centering

\includegraphics[width=0.6\linewidth]{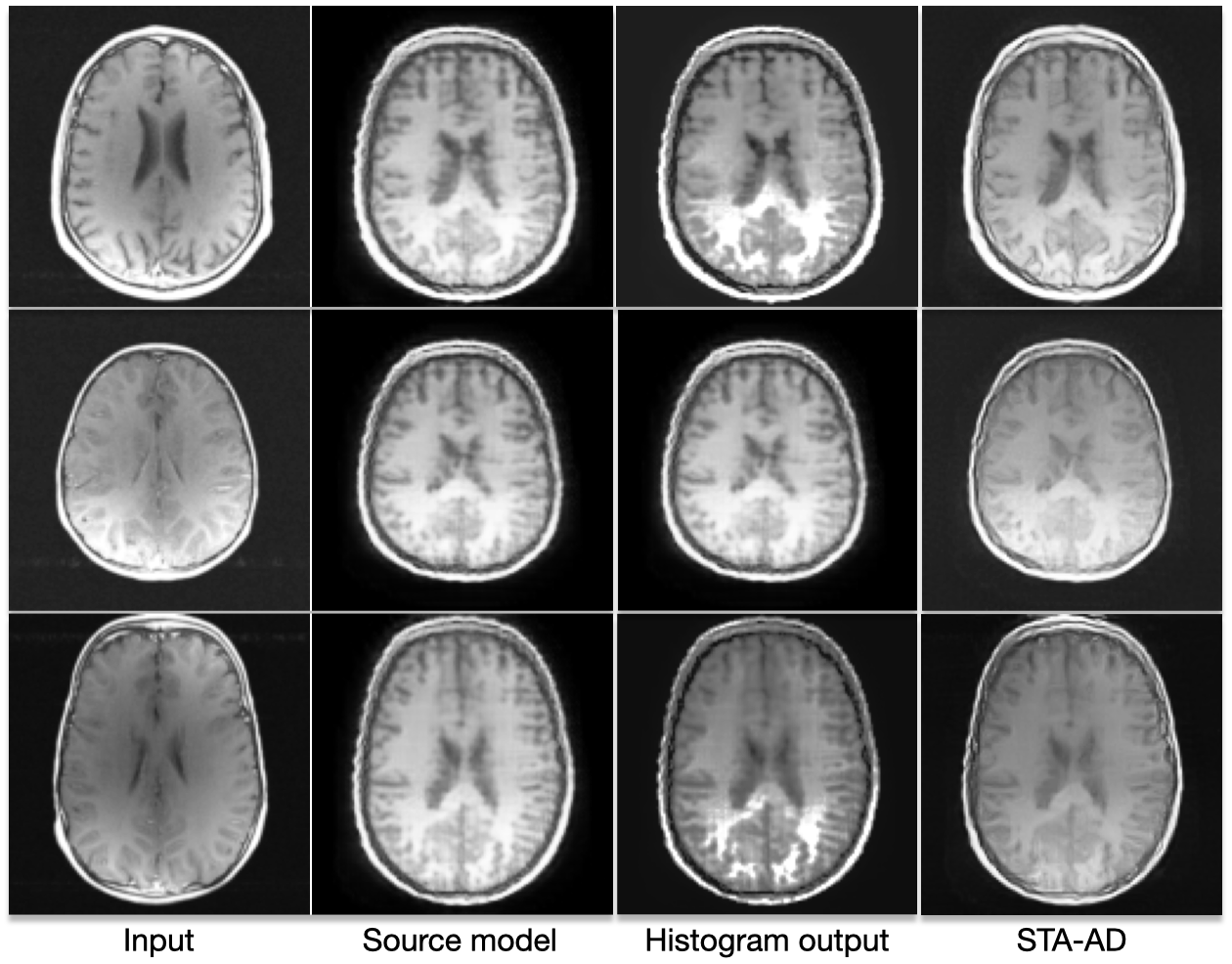}

\caption{\textbf{Visual results of histogram matching and our method on the target dataset. } These outputs are obtained by using the target input image and source model predictions as inputs. For histogram matching, the outputs result in visually incorrect results, introducing significant artifacts that degrade the final image quality. Whereas our method yields an adapted image that preserves image quality and thereby improves anomaly detection.  }
\label{fig1:hist_fig}

\end{figure*}

\end{document}